\documentclass{article} 
\usepackage{hyperref}
\usepackage{url}
\usepackage{verbatim}

\usepackage[top=1.2in, bottom=1.2in, left=1.2in, right=1.2in]{geometry}

\usepackage{amsthm,amsmath,amssymb}
\usepackage{latexsym}
\usepackage{graphicx}
\usepackage{subfigure}
\usepackage{algorithm}
\usepackage{algpseudocode} 

\usepackage{color}

\title{Enhanced Lasso Recovery on Graph}
\date{}

\author{
Xavier Bresson\thanks{Institute of Electrical Engineering, ({\tt xavier.bresson@epfl.ch})} \and
Thomas Laurent\thanks{Department of Mathematics, Loyola Marymount University ({\tt  tlaurent@lmu.edu})} \and
James von Brecht\thanks{Department of Mathematics, University of California Los Angeles ({\tt jub@math.ucla.edu})}  
}

\begin{document}

\maketitle
\begin{abstract}
This work aims at recovering signals that are sparse on graphs. {\it Compressed sensing} offers techniques for signal recovery from a few linear measurements and {\it graph Fourier analysis} provides a signal representation on graph. In this paper, we leverage these two frameworks to introduce a new Lasso recovery algorithm on graphs. More precisely, we present a {\it non-convex}, non-smooth algorithm that outperforms the standard convex Lasso technique. We carry out numerical experiments on three benchmark graph datasets.
\end{abstract}
%
%

\section{Sparse Representation on Graphs}
\label{sec1}

The goal of this work is to reconstruct signals on graphs that are supposed to be sparse in the graph Fourier representation. In this context, we will deal here with two main concepts, {\it graph} and {\it sparsity}, which have gathered a lot of attention in the recent years with the emergence of Compressed Sensing and Big Data. Let us introduce briefly these two concepts in the rest of this section.

\begin{figure}[ht]
\centering
\subfigure[LFR]{\includegraphics[width=4.5cm]{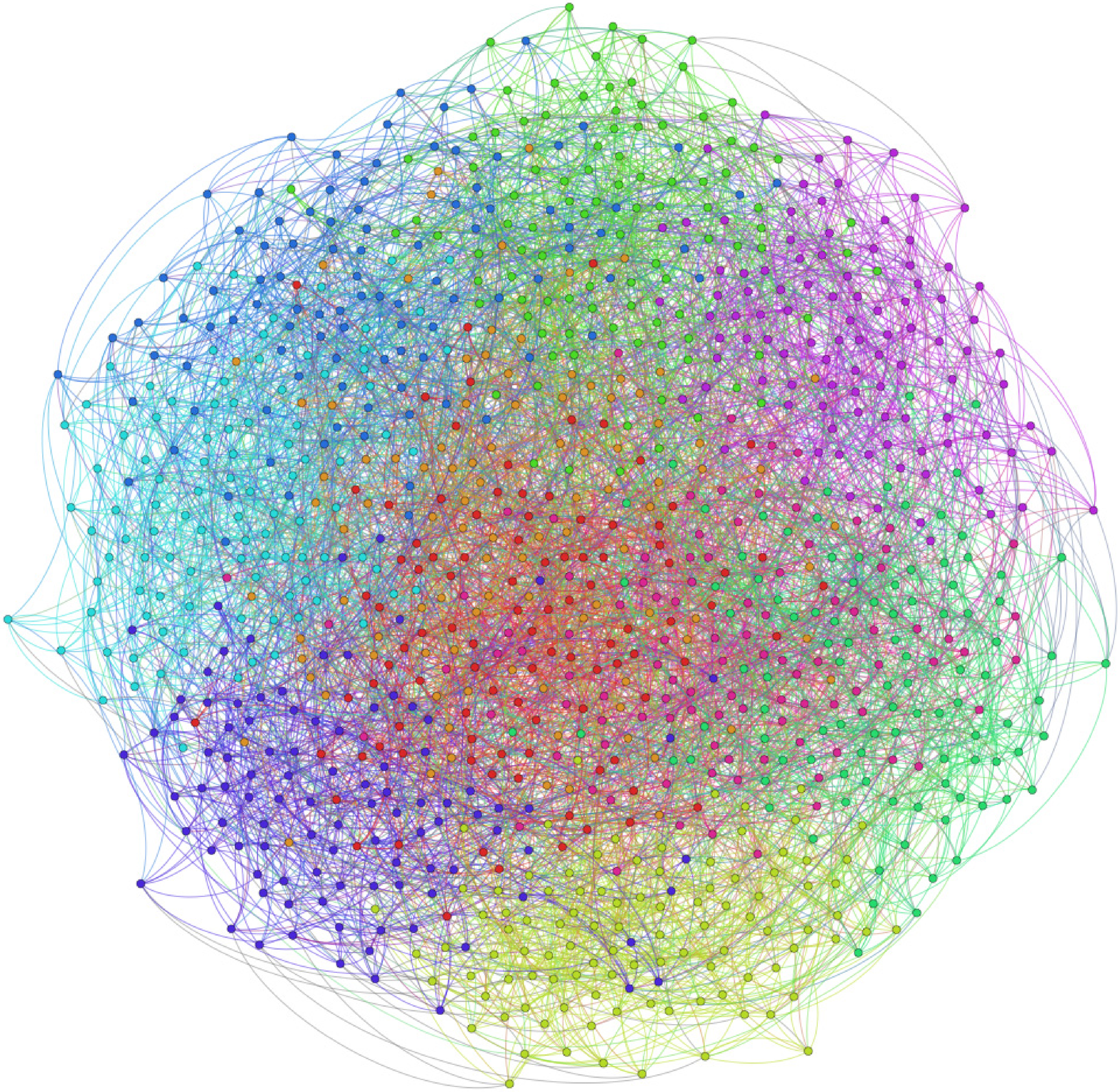}}
\hspace{0.2cm}
\subfigure[MNIST]{\includegraphics[width=4.5cm]{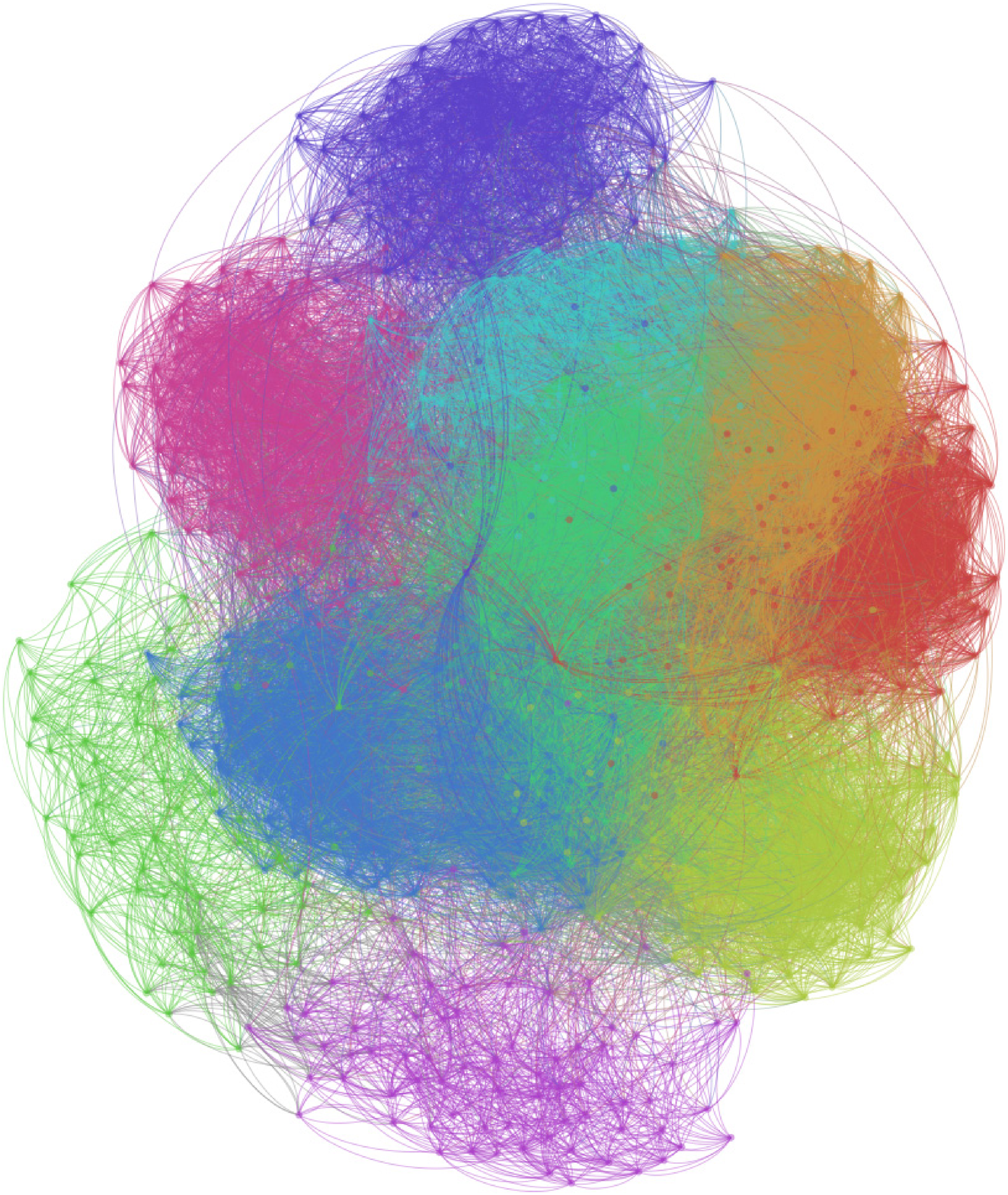}}
\hspace{0.2cm}
\subfigure[20NEWS]{\includegraphics[width=4.5cm]{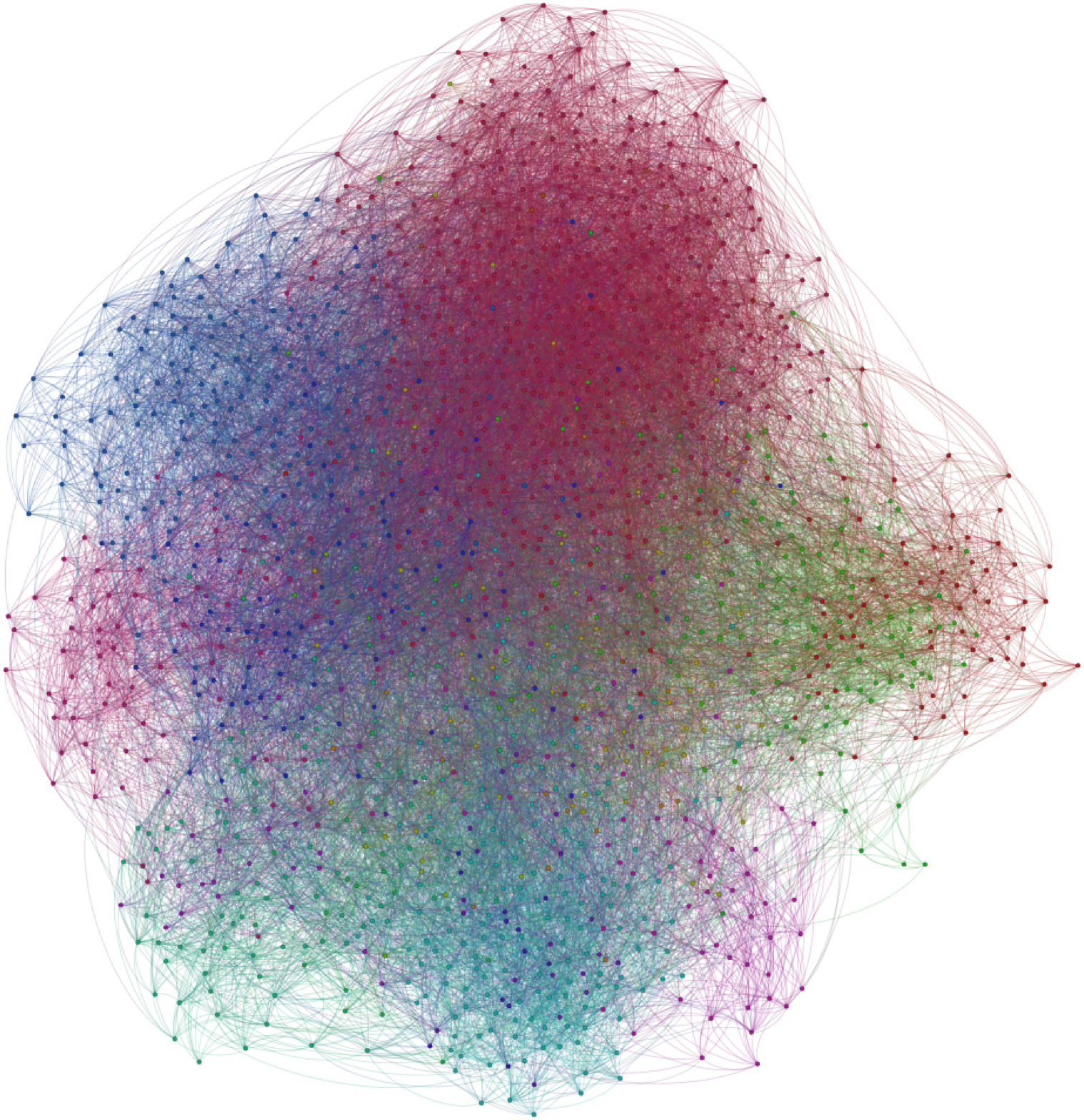}}\\
\subfigure[LFR]{\includegraphics[width=4.5cm]{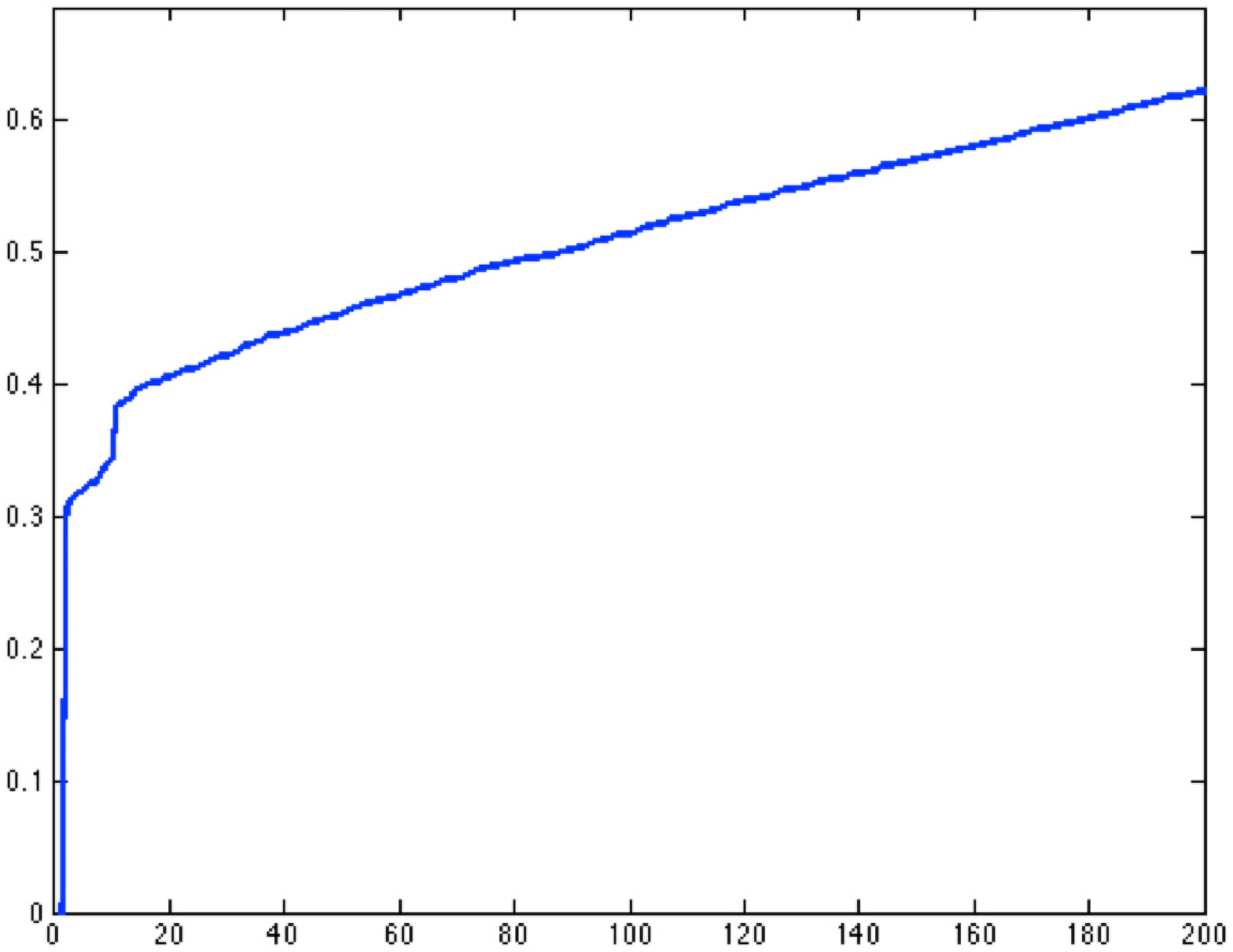}}
\hspace{0.2cm}
\subfigure[MNIST]{\includegraphics[width=4.5cm]{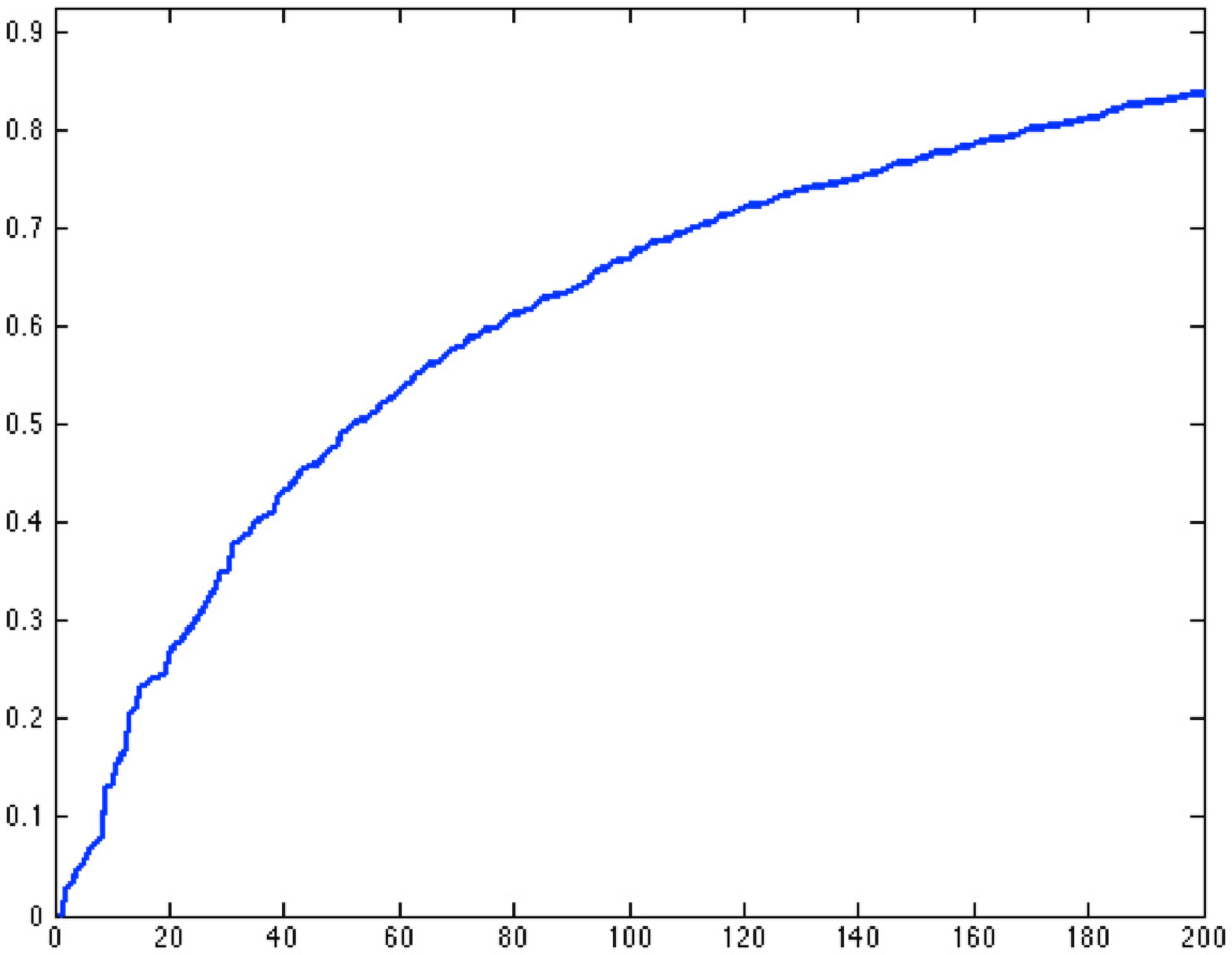}}
\hspace{0.2cm}
\subfigure[20NEWS]{\includegraphics[width=4.5cm]{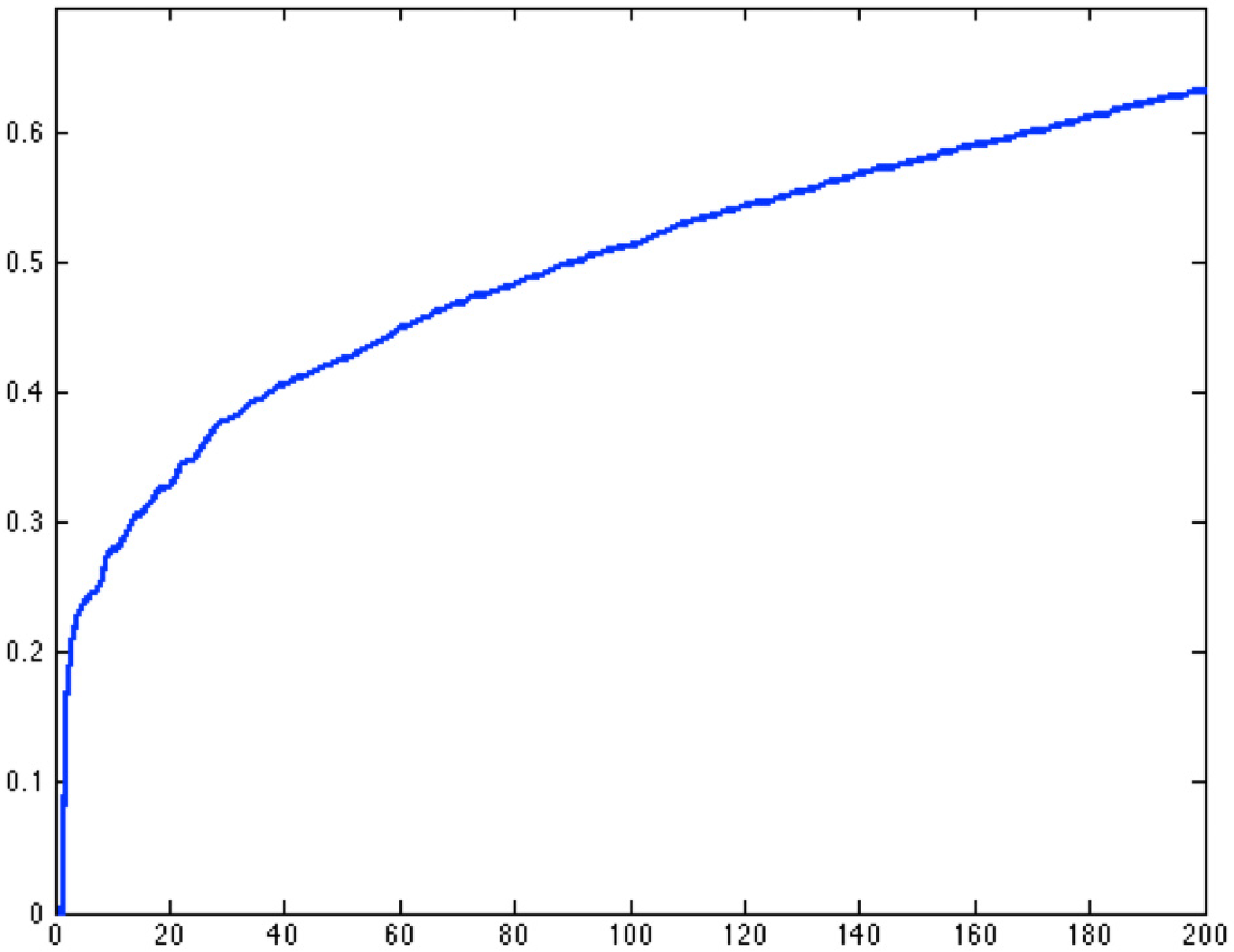}}
\caption{Graph and spectrum of LFR, MNIST, 20NEWS.}
\label{fig1}
\end{figure}

Graph/network is a powerful tool to represent complex high-dimensional datasets, in the sense that a graph structures data with respect to their similarities. Graphs have become increasingly more considered in applications such as search engines, social networks, airline routes, 3D geometric shapes, human brain connectivity, etc. Mathematics offer strong theoretical tools to analyze graphs with Harmonic Analysis and Spectral Graph Theory. An essential graph analysis tool is the graph Laplacian operator, which is the discrete approximation of the continuum Laplace-Beltrami operator for smooth manifolds. It is known that the eigenvectors of the Laplace-Beltrami operator provide a local parametrization of the manifold \cite{art:BelkinNiyogiSindhwani06SSTrans}. Equivalently, the eigenvectors of the graph Laplacian, also called graph Fourier modes, provides a representation of the graph. Given a graph with $(V,E,W)$, $V$, $E$ and $W$ being respectively the set of $n$ nodes, the set of edges and the similarity/adjacency matrix, then the (unnormalized) graph Laplacian operator is defined as
\begin{eqnarray}
L=D-W,
\nonumber
\end{eqnarray}
where $D$ is the diagonal degree matrix s.t. $D_{ii}=\sum_j W_{ij}$. $L$ is symmetric and positive-semidefinite, i.e. its eigenvalues $\lambda_i, \forall i$ are nonnegative. The graph Fourier modes are given by the eigenvectors $\{u_i\}_{i=1}^n$ of $L$ and can be represented by the orthogonal matrix $U=(u_1,...,u_n)\in\mathbb{R}^{n\times n}$ s.t. $U^\star U = I$. The graph Fourier basis $U$ acts as a basis to represent, analyze and process signals on graph. For example, one can represent a function $f:V\rightarrow \mathbb{R}$ on graph as $f(i)=\sum_{l=1}^n \hat{f}_l \cdot u_l(i)$ where $\hat{f}_l=\langle f,u_l \rangle=\sum_{i=1}^n f(i) \cdot u_l(i)$ is its Fourier transform. In this paper, we consider three well-known graphs. First, the synthetic LFR graph, which was introduced in \cite{art:LancichinettiFortunatoRadicchi08LFR} to study community graphs. Here, the number of nodes is chosen to be $n=1,000$, the number of communities is $10$ and the degree of community overlapping is $\mu=0.4$. Second, we consider a coarse version (for computational speedup) of the benchmark MNIST dataset of NYU \cite{pro:LeCunBottouBengioHaffner98MNIST} with $n=1,176$ nodes and the number of classes is $10$. Last, we use a coarse version of the well-known 20newsgroups dataset of CMU \cite{art:Joachims9620NEWS} with $n=1,432$ nodes and the number of classes is $20$. All three dataset graphs are illustrated on Figure \ref{fig1} with their graph Laplacian spectrum.

Sparse recovery is currently one of the most studied topics in signal processing. The main goal is to reconstruct signals that are supposed to be sparse in some basis representation. For example, in medical imaging, one of the objectives of sparsity is to speed up MRI acquisition by reconstructing an image in the Fourier basis given a small number of Fourier samples. This problem can be generalized to find the solution of a underestimated linear system of equations, which is generally ill-posed, with the constraint that the solution is sparse. Finding the solution of this problem is however impracticable because it is a NP-hard combinatorial problem. But Candes, Romberg, Tao and Donoho showed in \cite{art:CandesRombergTao06CS,art:Donoho06CS} that using an $\ell_1$ relaxation and under some conditions on the linear operator, known as the Restricted Isometry Property (RIP), and the measurements, known as incoherence property, there exists a tight convex relaxation of the NP-hard problem, that is easily tractable. However, it has recently been observed that the $\ell_1$ relaxation technique can be improved with reweighed $\ell_1$ \cite{art:CandesWakinBoyd08ReWeighL1}, $\ell_p, p<1$ \cite{art:Chartrand07NonConCS}, difference of convex functions $\ell_1$-$\ell_2$ \cite{art:LouOsherXin15DC} and smoothed $\ell_1/\ell_2$ ratio\cite{art:RepettiQuyenDuvalChouzenouxPesquet15L1L2}. These recent works suggest that non-convex relaxations may outperform the original $\ell_1$ sparse recovery. In this work, we follow this line of research and we introduce a new non-convex algorithm for sparse recovery on graph. Specifically, our goal is to improve Lasso problems on graph.

\section{Enhanced Sparsity}
\label{sec2}

\begin{figure}[ht]
\centering
\subfigure[$\ell_1$]{\includegraphics[width=5cm]{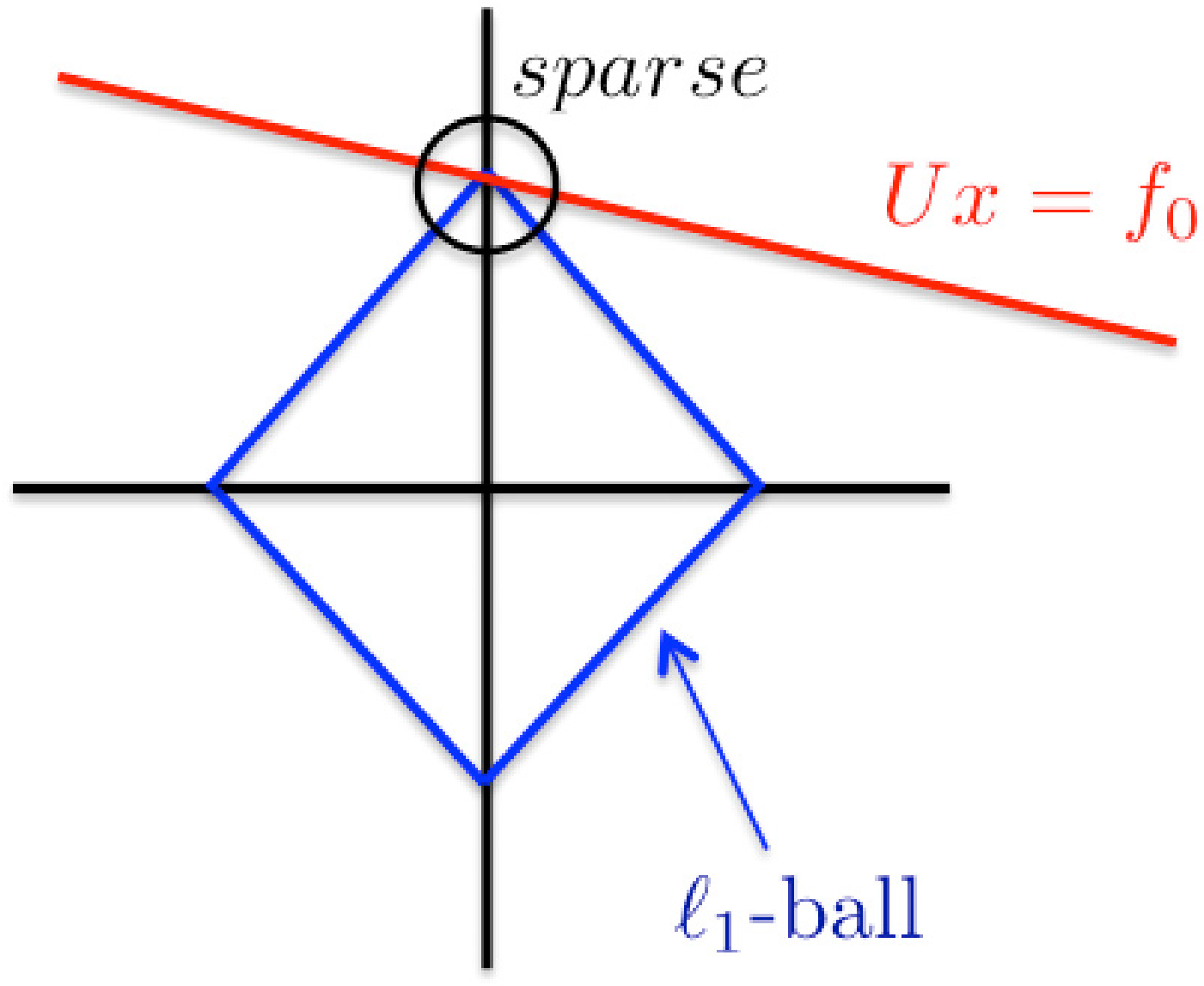}}
\hspace{0.2cm}
\subfigure[$\ell_1/ \ell_2$]{\includegraphics[width=5cm]{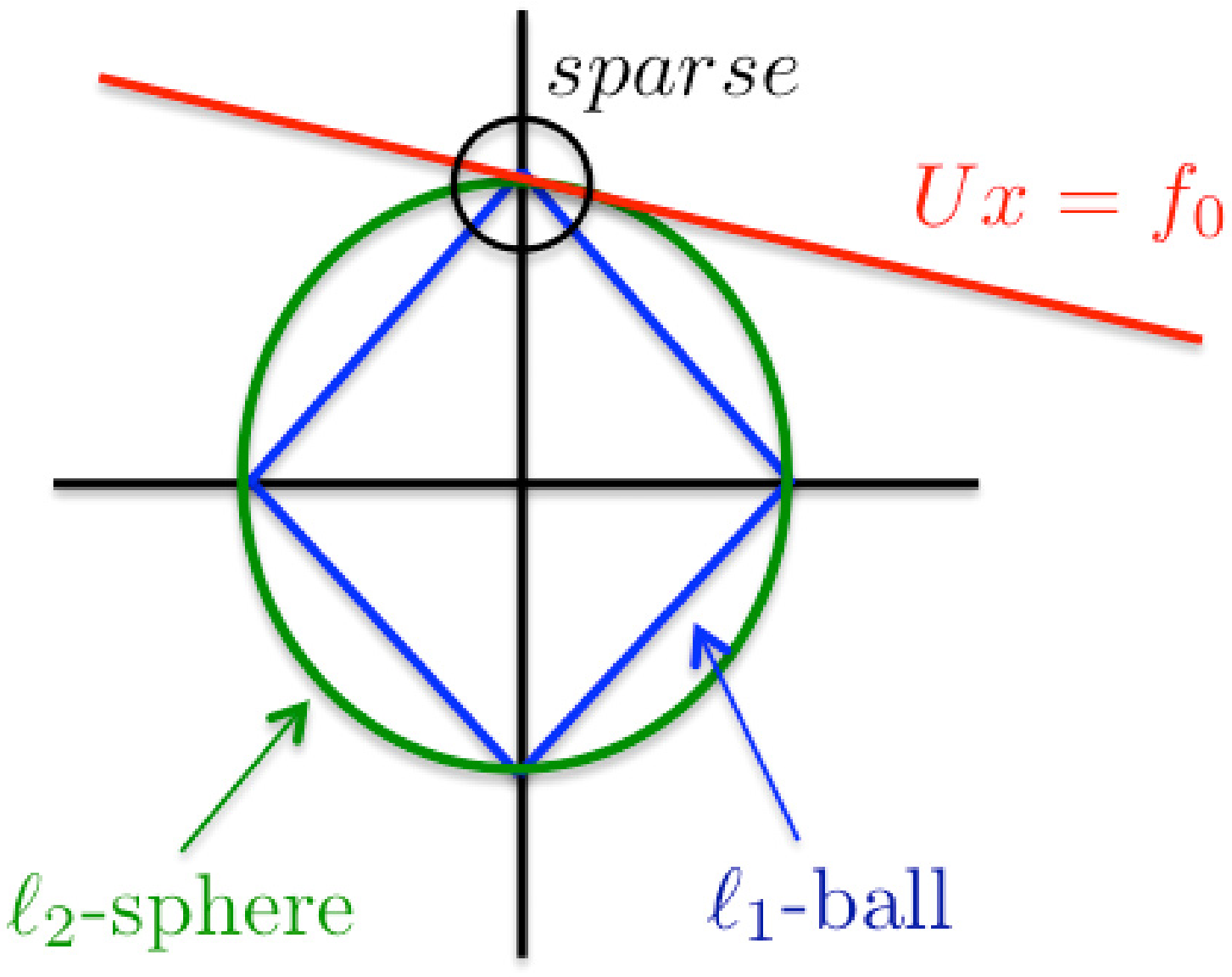}}
\caption{Standard $\ell_1$ and $\ell_1/ \ell_2$.}
\label{fig2}
\end{figure}

Starting from the standard $\ell_1$ problem for sparse recovery
\begin{eqnarray}
\min_x  \| x \|_1 \quad \textrm{ s.t. } \quad Ux=f_0,
\nonumber
\end{eqnarray}
where $x$ is a sparse signal to be recovered, $U$ is the graph Fourier basis, and $f_0$ are the given measurements, we propose the following enhanced recovery model
\begin{eqnarray}
\min_x  \| x \|_1 \quad \textrm{ s.t. } \quad Ux=f_0, \quad  \| x \|_2=1.
\nonumber
\end{eqnarray}
The new additional constraint, i.e. the {\it $\ell_2$ unit sphere}, is a non-convex set that is here essential for enhancing sparse recovery. Basically, it forces the solution to be at the intersection of the $\ell_1$-ball and the $\ell_2$-sphere, which are precisely the locations of sparse points in the Euclidean domain, see Figure \ref{fig2}. Observe now that the new  constrained $\ell_1$ optimization problem is equivalent to
 \begin{eqnarray}
\min_x \frac{ \| x \|_1 }{ \| x \|_2 } \quad \textrm{ s.t. } \quad Ux=f_0
\label{eq_ratio}
\end{eqnarray}
The equivalence comes from the fact that the ratio $\ell_1/ \ell_2$ is a {\it zero-homogenous} function, i.e. $F(\alpha x)=F(x), \alpha>0$. This means that the solution $x^\star$ is the same as $\alpha x^\star$, $\forall \alpha$. Particularly, for the specific value of $\alpha$ such that $x^\star$ belongs to the unit sphere $\| x^\star \|_2=1$. Figure \ref{fig2} compares geometrically the standard $\ell_1$ and the new ratio model $\ell_1/ \ell_2$. At a first glance, both models promote sparsity and the new model does not appear to bring anything new but a more complex problem. However, this figure acts as a simple illustration and one must remember that the recovery performance depends also on the incoherence property about the number of observed  measurements. In this context, the major motivation to go beyond convexity with the recent works \cite{art:CandesWakinBoyd08ReWeighL1,art:Chartrand07NonConCS,art:LouOsherXin15DC} is to precisely improve sparse recovery with a {\it smaller} number of measurements than the standard approach. We will see that the newly proposed model holds this property.

\section{Optimization}
\label{sec3}

We consider a different version of \eqref{eq_ratio} that is robust to noise:
\begin{eqnarray}
\min_x \frac{ \| x \|_1 }{ \| x \|_2 } + \frac{\lambda}{2}\| Ux-f_0 \|_2^2
\label{eq_ratio_robust}
\end{eqnarray}
Problem \eqref{eq_ratio_robust} is a non-smooth and non-convex optimization problem. The $\ell_1$/non-smooth part of the problem can be handled quite efficiently with techniques introduced in Compressed Sensing such as Alternating Direction Method of Multipliers (ADMM) \cite{art:GoldsteinOsher09SB} or Uzawa-type Primal-Dual technique \cite{art:ZhangBurgerBressonOsher10SparseRepre}. However, the non-convex part is more challenging. For general non-convex problems, it is difficult to design an algorithm that is fast, accurate, robust and also guaranteed to converge, or at least that satisfies the monotonicity property. Monotonicity means that the energy is guaranteed to decrease at each iteration, although the problem is non-convex. In this situation, most non-convex algorithms only find solutions that are critical points or local minimizers, and rarely global minimizers.

\subsection{Proximal Forward-Backard Splitting Algorithm}
\label{sec3a}

We develop in this section an algorithm for the ratio optimization problem \eqref{eq_ratio_robust}. A related numerical scheme was introduced in \cite{art:BressonLaurentUminskyVonBrecht13MTV} in the different context of data clustering. Let $T(x)=\| x \|_1$, $B(x)=\| x \|_2$, $E(x)=T(x)/B(x)$ and $F(x)=\frac{\lambda}{2}\| Ux-f_0 \|_2^2$ such that we want to solve 
\begin{eqnarray}
\min_x \frac{ T(x) }{ B(x) } + F(x).
\nonumber
\end{eqnarray}
Let us consider a semi-implicit gradient flow for this problem: 
\begin{eqnarray}
\frac{x^{k+1}-x^k}{\tau^k}&=&-\frac{ \partial T(x^{k+1})\cdot B(x^k) - T(x^k) \cdot \partial B(x^k) }{ B^2(x^k) }  - \partial F(x^{k+1}),\nonumber
\end{eqnarray}
where $\partial$ stands for the subdifferentials of $T$ and $B$ (which is not unique for $\ell_1$ but is for $\ell_2$) and $\tau^k$ is the time step. This provides the optimality condition
\begin{eqnarray}
x^{k+1} - (x^k+\tau^k\frac{E^k}{B^k} \partial B(x^k)) + \frac{\tau^k}{B^k}\partial T^{k+1} +\tau^k \partial F^{k+1} \ni 0,
\label{eq_opt_cond}
\end{eqnarray}
where the notations $T^k=T(x^k)$ and $B^k=B(x^k)$ are used. This leads to a two-step iterative scheme:\\
\begin{eqnarray}
(1)\quad y^k = x^k+ c_0^k  \partial B(x^k) 
\nonumber
\end{eqnarray}
and
\begin{eqnarray}
(2)\quad x^{k+1}&=&\arg\min_x c_1^k T(x) + \frac{\tau^k}{2} F(x) + \frac{1}{2} \| x-y^k\|_2^2 \nonumber \\
&=&\textrm{prox}_{c_1^k T+\frac{\tau^k}{2} F}(y^k),
\nonumber
\end{eqnarray}
where $c_0^k=\tau^kE^k/B^k$ and $c_1^k=\tau^k/B^k$. The second step is the proximal operator \cite{art:ZhangBurgerBressonOsher10SparseRepre,art:ChambollePock11FastPD} of the convex function $c_1^k T+\frac{\tau^k}{2} F$. Overall, we have designed a proximal forward-backward splitting algorithm to solve \eqref{eq_ratio_robust} as the solution is given by
\begin{eqnarray}
x^{k+1}=\textrm{prox}_{c_1^k T+\frac{\tau^k}{2} F}( x^k+ c_0^k  \partial B(x^k) ).
\label{eq_prox_algo}
\end{eqnarray}
In the next section, we will show that the proposed iterative algorithm is (almost) monotonic, i.e. its energy is guaranteed to decrease at each iteration.

\subsection{Monotonicity}
\label{sec3b}

We show the following quasi-monotonicity result:
\begin{eqnarray}
\frac{B^{k+1}}{B^k}(E^k-E^{k+1}) + (F^k-F^{k+1})\geq \frac{\| x^k-x^{k+1}\|_2^2}{\tau^k}
\label{eq_mono}
\end{eqnarray}

\noindent
{\it Proof.} Define the convex functions
\begin{eqnarray}
\mathcal{G}^k(x)&=&c_0^k B(x)+\tau^k F(x), \label{eq_defF} \\
\mathcal{F}^k(x)&=&c_1^kT(x)+\tau^k F^k, \label{eq_defG}
\end{eqnarray}
and observe that $\mathcal{G}^k(x^k)=\mathcal{F}^k(x^k)$ for latter use. We remind the general definition of the subdifferential $ \partial \mathcal{E}$ of a convex function $\mathcal{E}$:
\begin{eqnarray}
\mathcal{E}(x_1) \geq \mathcal{E}(x_2) + \langle x_1 - x_2 , y_2 \rangle, \ \forall y_2 \in \partial \mathcal{E}(x_2).
\label{eq_subdiff}
\end{eqnarray}
We plug $x_1=x^{k+1}$, $x_2=x^k$ and $\mathcal{E}=\mathcal{G}$ in \eqref{eq_subdiff}:
\begin{eqnarray}
\mathcal{G}^k(x^{k+1}) \geq \mathcal{G}^k(x^k)  + \langle x^{k+1} - x^k ,\partial  \mathcal{G}^k(x^k) \rangle
\label{eq_G1}
\end{eqnarray}
If we now observe that the first step of the algorithm is $y^k=x^k+v^k$ with $v^k=c_0^k \partial B(x^k)=\partial \mathcal{G}^k(x^k)$ then \eqref{eq_G1} becomes
\begin{eqnarray}
\mathcal{G}^k(x^{k+1}) \geq \mathcal{G}^k(x^k)  + \langle x^{k+1} - x^k ,v^k \rangle.
\label{eq_G2}
\end{eqnarray}
Let us now plug $x_1=x^k$, $x_2=x^{k+1}$ and $\mathcal{E}=\mathcal{F}$ in \eqref{eq_subdiff}:
\begin{eqnarray}
\mathcal{F}^k(x^k) \geq \mathcal{F}^k(x^{k+1})  + \langle  x^k - x^{k+1} ,\partial  \mathcal{F}^k(x^{k+1}) \rangle.
\label{eq_F1}
\end{eqnarray}
Notice that the optimality condition \eqref{eq_opt_cond} reads $x^{k+1}-y^k + \partial  \mathcal{F}^k(x^{k+1}) \ni 0$ and thus $y^k-x^{k+1}\in \partial \mathcal{F}^k(x^{k+1})$. This implies that \eqref{eq_F1} may be written as
\begin{eqnarray}
\mathcal{F}^k(x^k) &\geq& \mathcal{F}^k(x^{k+1})  + \langle  x^k - x^{k+1} ,y^k-x^{k+1} \rangle \nonumber \\
&\geq& \mathcal{F}^k(x^{k+1})  + \| x^k-x^{k+1}\|_2^2 + \langle x^k - x^{k+1} ,v^k \rangle 
\label{eq_F2}
\end{eqnarray}
Adding \eqref{eq_G2} and \eqref{eq_F2} and using the fact that $\mathcal{G}^k(x^k)=\mathcal{F}^k(x^k)$ we have
\begin{eqnarray}
\mathcal{G}^k(x^{k+1}) \geq \mathcal{F}^k(x^{k+1})  + \| x^k-x^{k+1}\|_2^2
\label{eq_FG}
\end{eqnarray}
Using the definition \eqref{eq_defF} and \eqref{eq_defG}, this inequality can be rewritten as \eqref{eq_mono}, which is the desired result. $\Box$\\


\noindent
{\it Notes.} Observe that close to the steady-state solution, we have $B^{k+1}/B^k\rightarrow 1 $ for $k\rightarrow \infty$ and the quasi-monotonicity tends to a monotonicity property. Second, see that if we had access to the quantity $B^{k+1}$ (or a good estimation) then we would set $\tau^k=\frac{B^k}{B^{k+1}}\tau_0$ and this would imply
\begin{eqnarray}
E_{Tot}^k-E_{Tot}^{k+1}\geq \| x^k-x^{k+1}\|_2^2/\tau^k,
\nonumber
\end{eqnarray}
where $E_{Tot}=E+F$, and thus unconditional monotonicity for any $\tau_0$.

\section{Applications}
\label{sec4}

\subsection{Enhanced Lasso on Graphs}
\label{sec4a}

{\bf The Algorithm.} The standard Lasso problem on graph is $\min_x  \| x \|_1 + \frac{\lambda}{2}\| Ux-f_0 \|_2^2$ where $U$ is the sensing matrix, here the graph Fourier modes. Function $f_0$ is the signal measured on the graph. It is generated as $f_0=U(x_0+n)$ where $x_0$ is a pure sparse signal with 5\% of non-zero entries uniformly chosen between $[-1,1]$ and $n$ is the noise, a Gaussian distribution with standard deviation $\sigma=0.1$. The goal is to recover the sparse signal $x_0$. We recall that the proposed enhanced Lasso problem on graph is $\min_x  \frac{ \| x \|_1 }{ \| x \|_2 } + \frac{\lambda}{2}\| Ux-f_0 \|_2^2$. We use the proximal forward-backward splitting algorithm introduced in Section \ref{sec3a} to solve it. That is, Step 1: $y^k = x^k+ \frac{\tau^k E^k}{B^k}  \partial  \| x \|_2 |_{x^k} = x^k+ \frac{\tau^k E^k}{B^k} \frac{x^k}{\| x^k \|_2} $, and Step 2: $x^{k+1}=\arg\min_x F(x) + G(x)$ where $F(x)=\| x \|_1$  and $G(x)=\frac{E^k \lambda}{2} \| Ux-f_0 \|_2^2 + \frac{E^k}{2\tau^k}  \| x-y^k\|_2^2$. We may write this problem as a saddle-point problem $\min_x \max_p \langle p , x \rangle - F^\star(p) + G(x)$ where $F^\star$ is the barrier function of the $\ell_\infty$ unit ball such that
\begin{eqnarray}
F^\star(p) = 
\left\{
\begin{array}{ll}
0 & \textrm{ if } |p|\leq 1,\\
+\infty & \textrm{ otherwise, }
\end{array}
\right.
\nonumber
\end{eqnarray}
Note that $G(x)$ is uniformly convex so that we can apply the accelerated primal-dual algorithm of \cite{art:ChambollePock11FastPD}. The algorithm consists in iterating the following steps:
\begin{eqnarray}
p^{n+1} &=& \textrm{prox}_{\sigma^n F^\star} (p^n + \sigma^n \bar{x}^n)\\
x^{n+1} &=& \textrm{prox}_{\eta^n G} (x^n - \eta^n p^{n+1})\\
\theta^{n+1} &=& 1/\sqrt{1+2\gamma\eta^n}, \tau^{n+1}=\theta^{n+1}\eta^n,  \sigma^{n+1}=\sigma^n/\theta^{n+1}\\
\bar{x}^{n+1} &=& x^{n+1} + \theta^{n+1} ( x^{n+1} - x^n )
\label{eq_pd}
\end{eqnarray}
The scheme converges quickly, with order $O(1/n^2)$, provided that $\sigma^0=\eta^0=1$. The first inner proximal problem has an analytical solution
\begin{eqnarray}
\textrm{prox}_{\sigma^n F^\star}(z)=z / \max \{ 1, |z| \},
\nonumber
\end{eqnarray}
and the second inner proximal problem has also a closed-form solution
\begin{eqnarray}
\textrm{prox}_{\eta^n G}(z) = \frac{ z + E^k\lambda \eta^n U^*f_0 + E^k\eta^n y^k/\tau^k  }{ 1 + E^k\lambda \eta^n + E^k\eta^n/\tau^k  }.
\nonumber
\end{eqnarray}
As the two proximal operators are  fast to solve, so it is for the general algorithm. In fact, {\it solving the non-convex ratio problem \eqref{eq_ratio_robust} for sparse recovery can be seen as solving the standard Lasso problem with the addition of a convex quadratic term $\| x-y^k\|_2^2$ and updating $y^k$ each time the monotonicity condition \eqref{eq_mono} is satisfied}. We summarize the algorithm here.\\

\noindent
{\bf Algorithm.} Initialize $x^0=U^*f_0$, $\sigma^{n=0}=\eta^{n=0}=1$, $\gamma=1$, and iterate $k$ until convergence \\
(1) $\tau^k  = B^k$\\
(2) $y^k = x^k+ E^k \frac{x^k}{\| x^k \|_2} $\\
(3) Inner loop: iterate $n$ until the monotonicity condition, $B^{n}/B^k(E^k-E^{n}) + (F^k-F^{n})\geq \| x^k-x^{n}\|_2^2/\tau^k$, is satisfied:\\
\indent (3i) $p^{n+1} = (p^n + \sigma^n \bar{x}^n) / \max \{ 1, |p^n + \sigma^n \bar{x}^n| \} $\\
\indent (3ii) $x^{n+1} = \frac{ x^n - \eta^n p^{n+1} + E^k\lambda \eta^n U^*f_0 + E^k\eta^n y^k/\tau^k }{1 + E^k\lambda \eta^n + E^k\eta^n/\tau^k }$\\
\indent (3iii) $\theta^{n+1} = 1/\sqrt{1+2\gamma\eta^n}, \tau^{n+1}=\theta^{n+1}\eta^n,  \sigma^{n+1}=\sigma^n/\theta^{n+1}$\\
\indent (3iv) $\bar{x}^{n+1} = x^{n+1} + \theta^{n+1} ( x^{n+1} - x^n )$\\
(4) $x^k = x^{n+1}$\\
{\it Note:} the time step $\tau^k  = B^k$ was chosen experimentally, and is the subject of future study.\\

\begin{figure}[h!]
\centering
\subfigure[LFR, $\ell_1$]{\includegraphics[width=4.5cm]{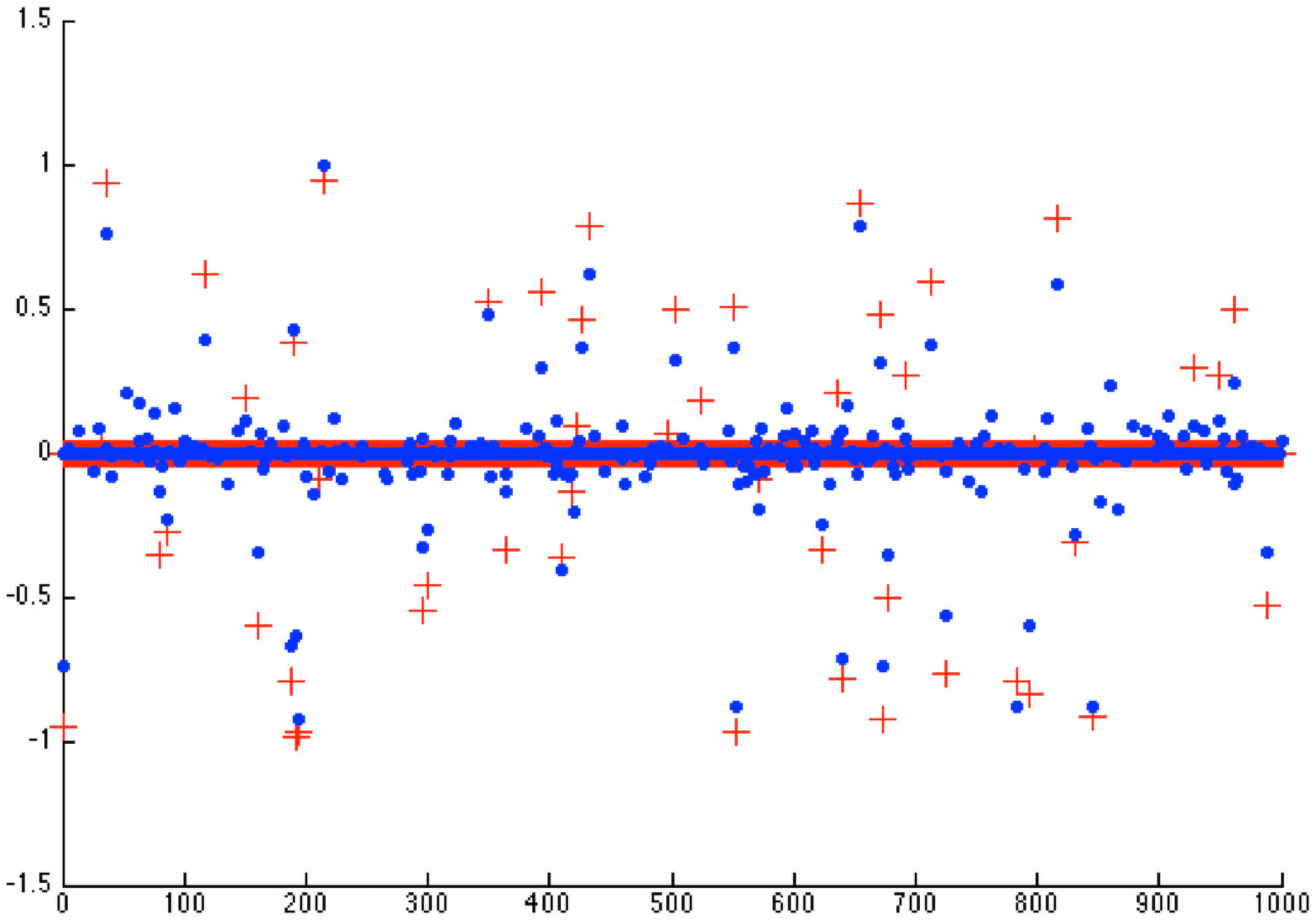}}
\hspace{0.2cm}
\subfigure[MNIST, $\ell_1$]{\includegraphics[width=4.5cm]{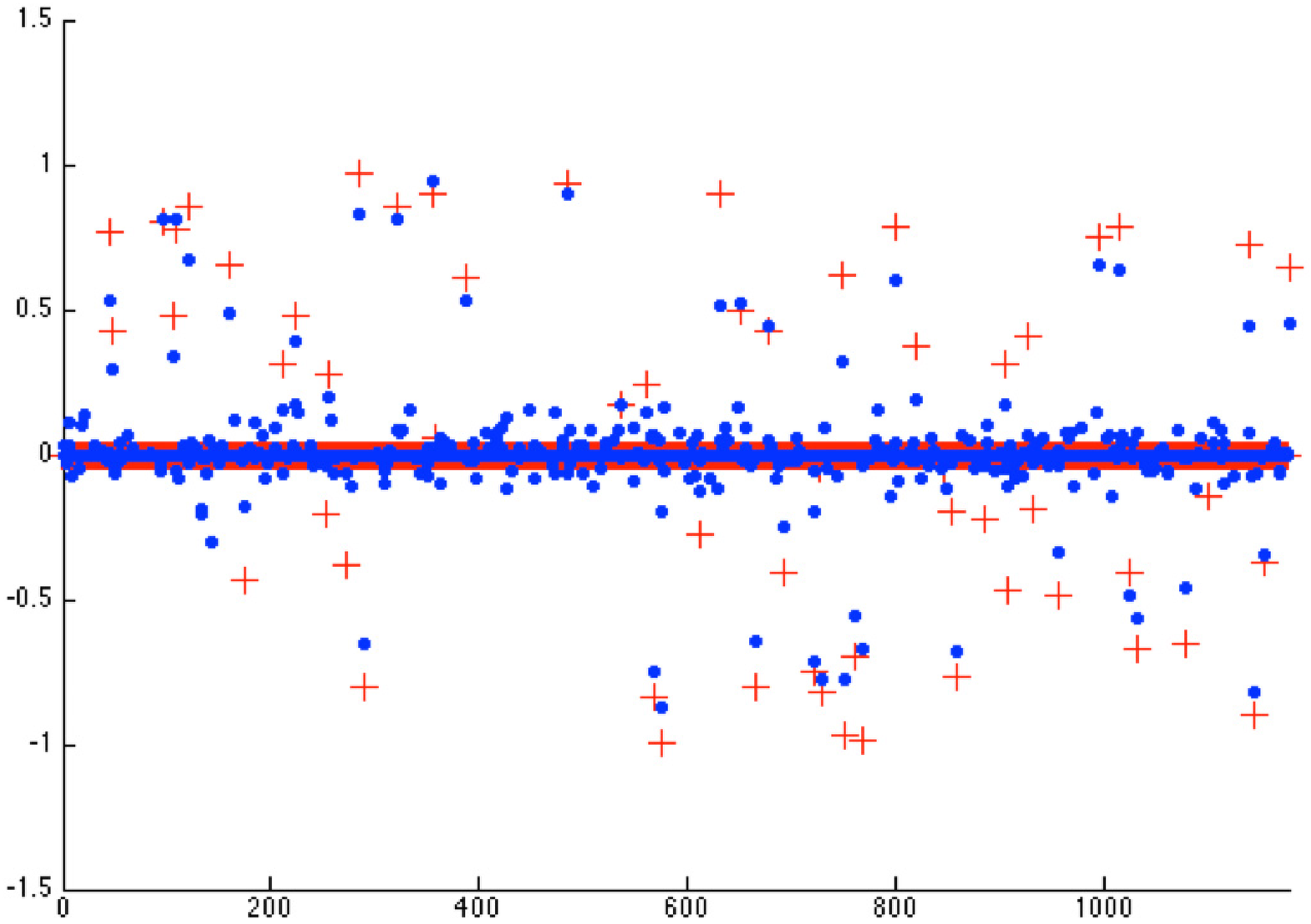}}
\hspace{0.2cm}
\subfigure[20NEWS, $\ell_1$]{\includegraphics[width=4.5cm]{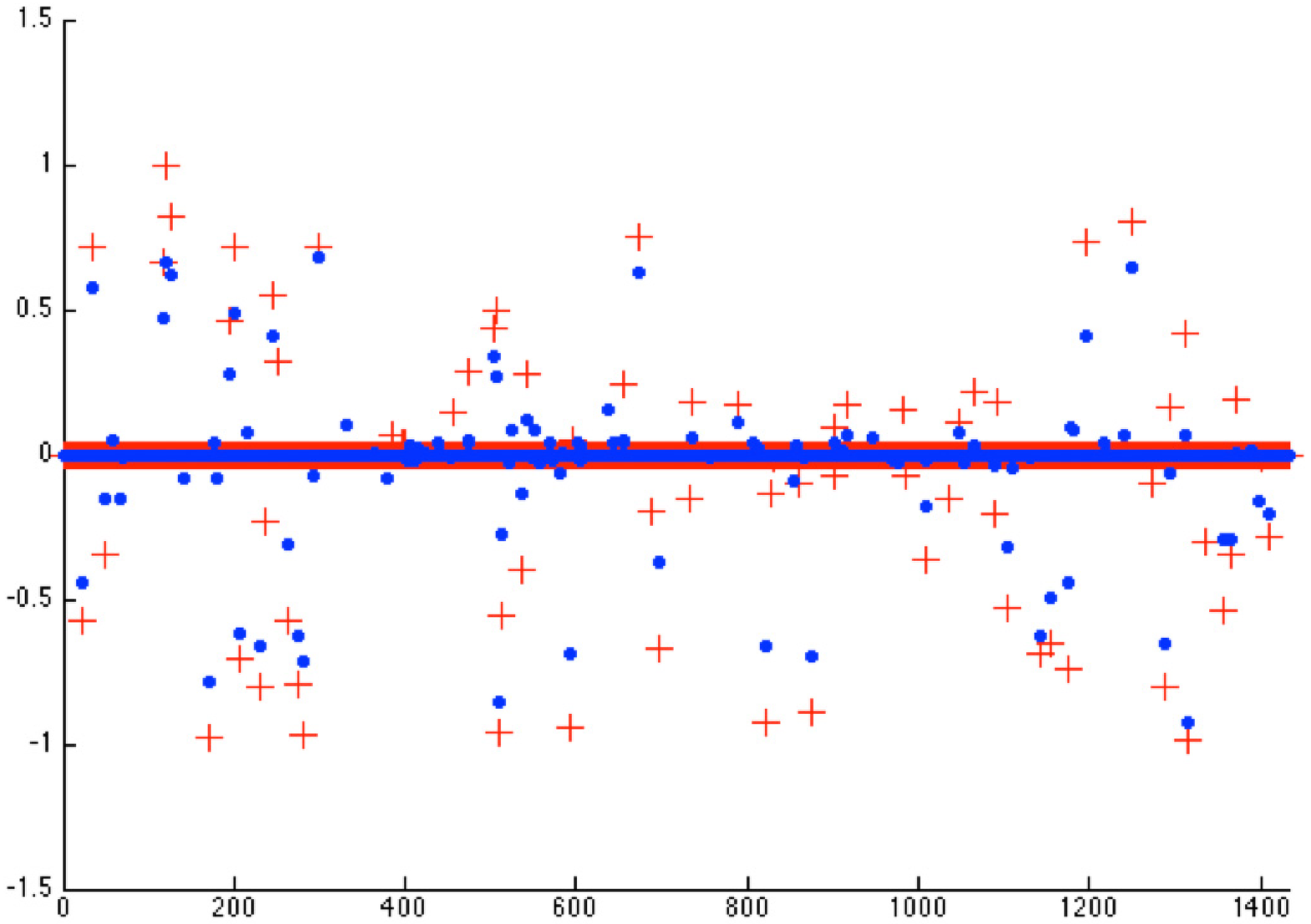}}\\
\subfigure[LFR, $\ell_1/ \ell_2$]{\includegraphics[width=4.5cm]{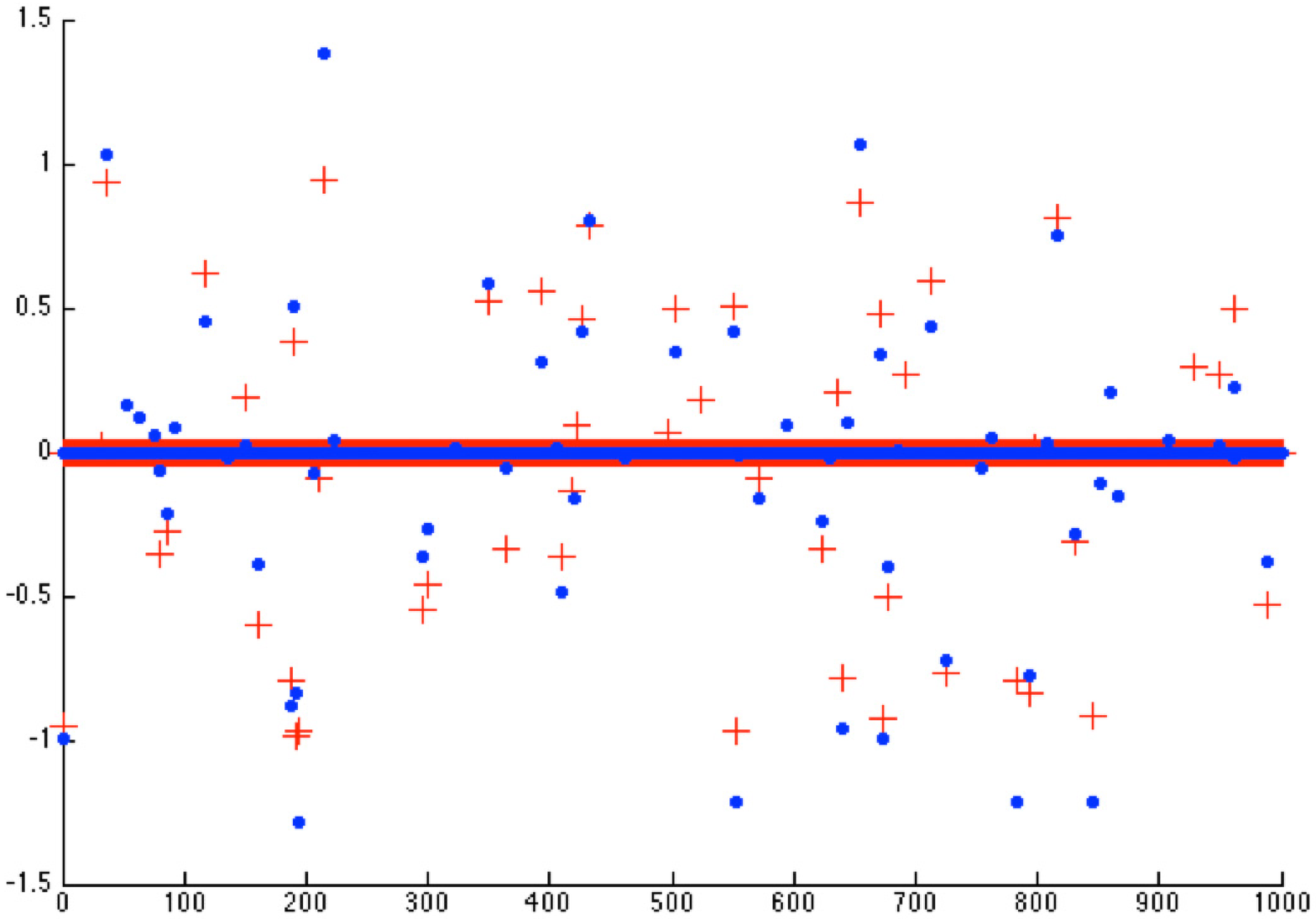}}
\hspace{0.2cm}
\subfigure[MNIST, $\ell_1/ \ell_2$]{\includegraphics[width=4.5cm]{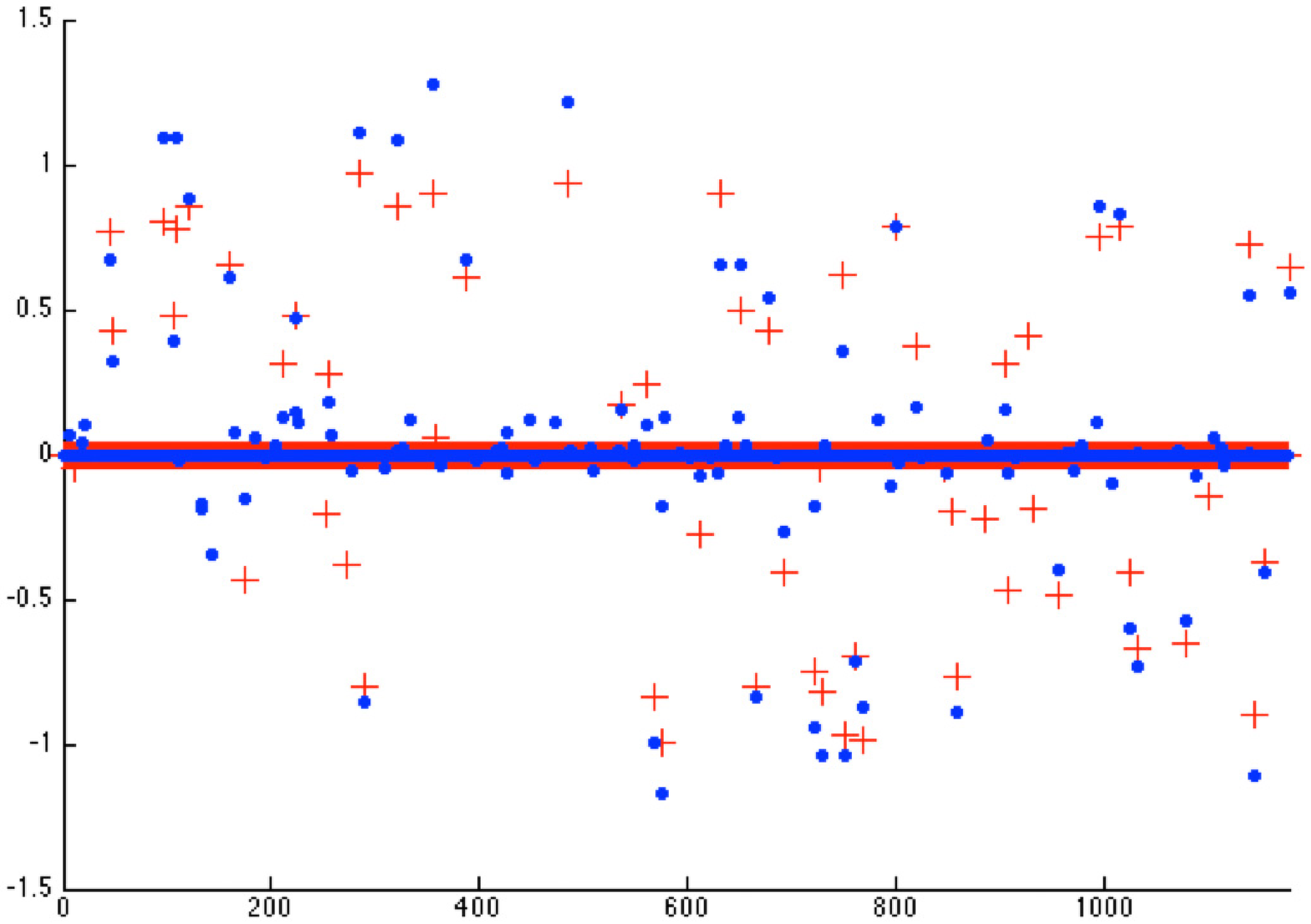}}
\hspace{0.2cm}
\subfigure[20NEWS, $\ell_1/ \ell_2$]{\includegraphics[width=4.5cm]{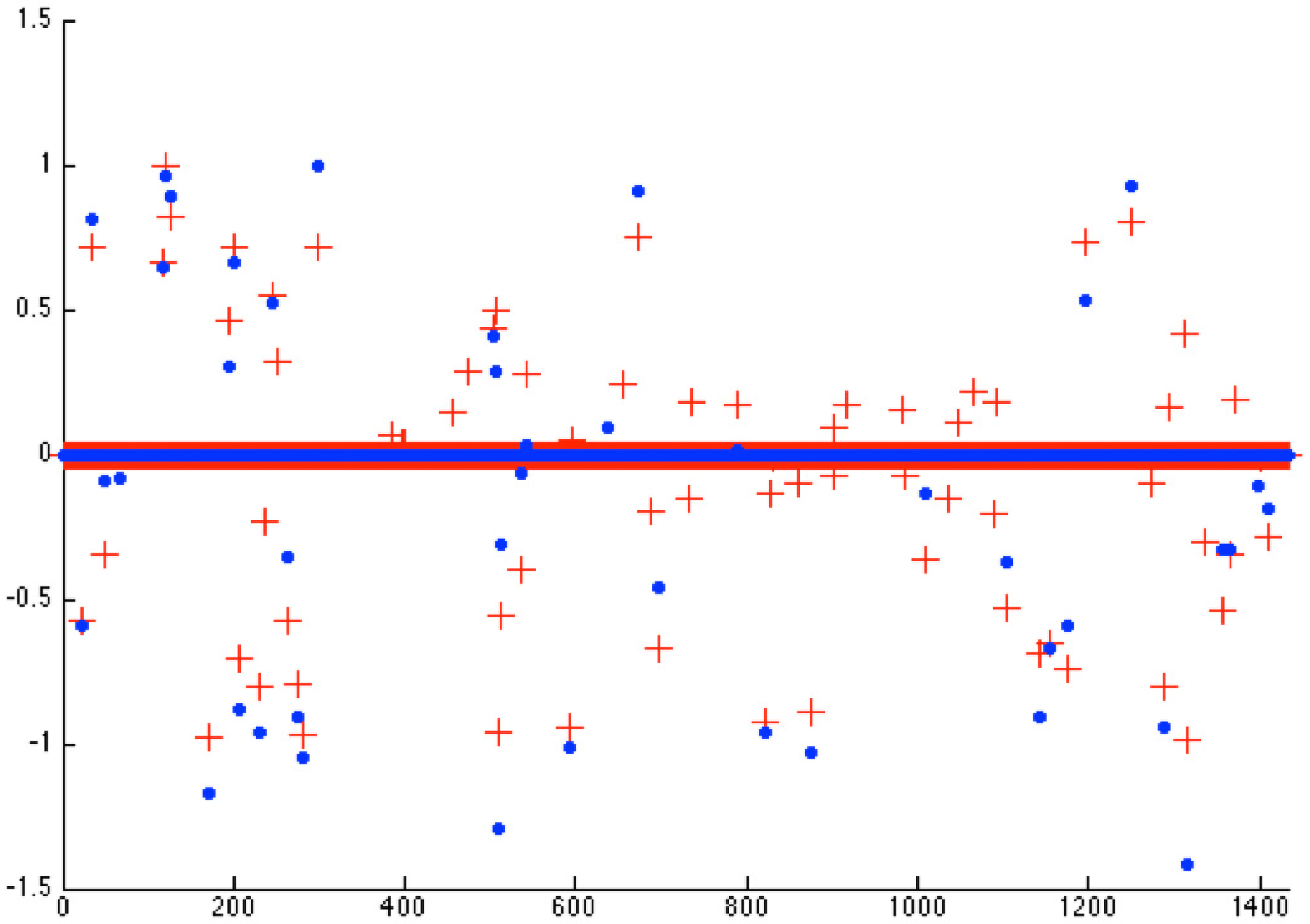}}
\caption{Standard Lasso vs Proposed Lasso on three graphs.}
\label{fig3}
\end{figure}

\noindent
{\bf Numerical Experiments.} We compare standard Lasso and enhanced Lasso on graphs. We test on the LFR, MNIST and 20NEWS graphs. The value of the parameter $\lambda$ that balances the sparsity term and the fidelity term is chosen to minimize the recovery error defined as $\| x-x_0\|_2 / \| x\|_2$ for all models and all graphs. The results are reported on Table \ref{tab1} and Figure \ref{fig3}. Overall, the proposed enhanced Lasso model performs better than the standard one, but it is 2-3 times slower.

\begin{table}[h!]
\centering
\begin{tabular}{|c|c|c|}
\hline
  & Standard Lasso  & Proposed Lasso \\
  \hline
 LFR & 0.419  & {\bf 0.309} \\
 MNIST & 0.417  & {\bf 0.302} \\
 20NEWS & 0.481  & {\bf 0.325} \\
 \hline	
\end{tabular}
\caption{Accuracy for standard Lasso vs proposed Lasso on three graphs.}
\label{tab1}
\end{table}

\subsection{Enhanced Lasso-Inpaiting on Graphs}
\label{sec4b}

{\bf The Algorithm.} In this section, we add a layer of difficulty by removing a set of observed measurements in $f_0$. In other words, we do not observe the whole function $f_0$ but only a portion of it. This problem is equivalent to a Lasso-Inpainting problem. For this, a diagonal selector matrix $R$ is added to the linear operator $U$ such that 
\begin{eqnarray}
R_{ii} = 
\left\{
\begin{array}{ll}
1 & \textrm{ if } i\in\Omega_{obs},\\
0 & \textrm{ otherwise, }
\end{array}
\right.
\nonumber
\end{eqnarray}
$\Omega_{obs}$ being the set of observed measurements, and $R_{ii}=0$ otherwise. The formulation is thus $\min_x  \| x \|_1 + \frac{\lambda}{2}\| RUx-f_0 \|_2^2$. The enhanced Lasso-Inpainting is naturally 
\begin{eqnarray}
\min_x \frac{ \| x \|_1 }{ \| x \|_2 }  + \frac{\lambda}{2}\| RUx-f_0 \|_2^2.
\nonumber
\end{eqnarray}
We apply the same technique as in Section \ref{sec4a} to compute a solution to the problem. The only change is the solution of the inner proximal problem $\textrm{prox}_{\eta^n G} (z) = U^* ( Ub/K )$ where $b=z+ E^k\lambda \eta^n RU^*f_0 + E^k\eta^n y^k/\tau^k$ and $K= I +E^k\lambda \eta^nR + E^k\eta^n/\tau^k$, which is also fast to compute. \\

\begin{figure}[h!]
\centering
\subfigure[LFR, $\ell_1$]{\includegraphics[width=4.5cm]{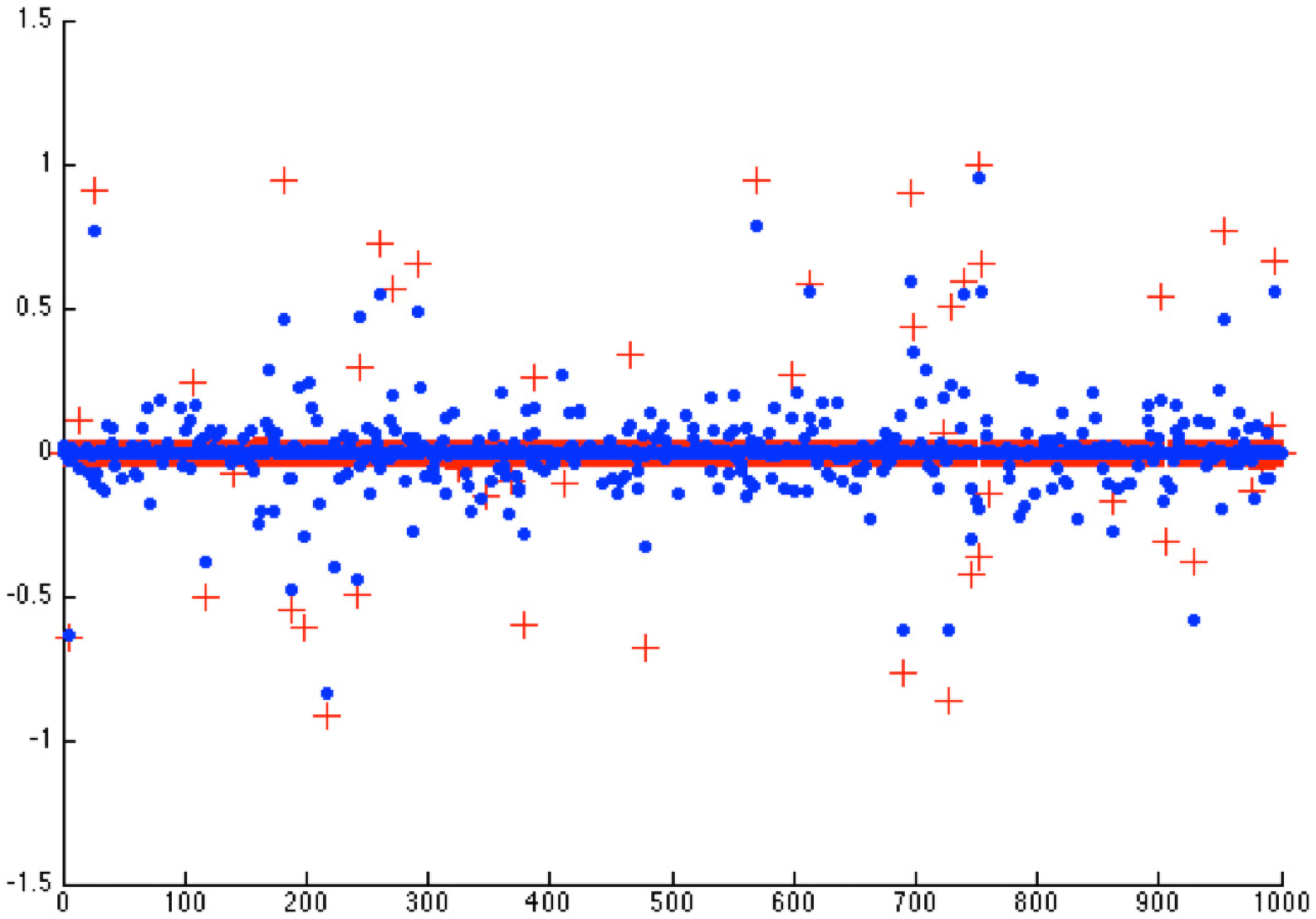}}
\hspace{0.2cm}
\subfigure[MNIST, $\ell_1$]{\includegraphics[width=4.5cm]{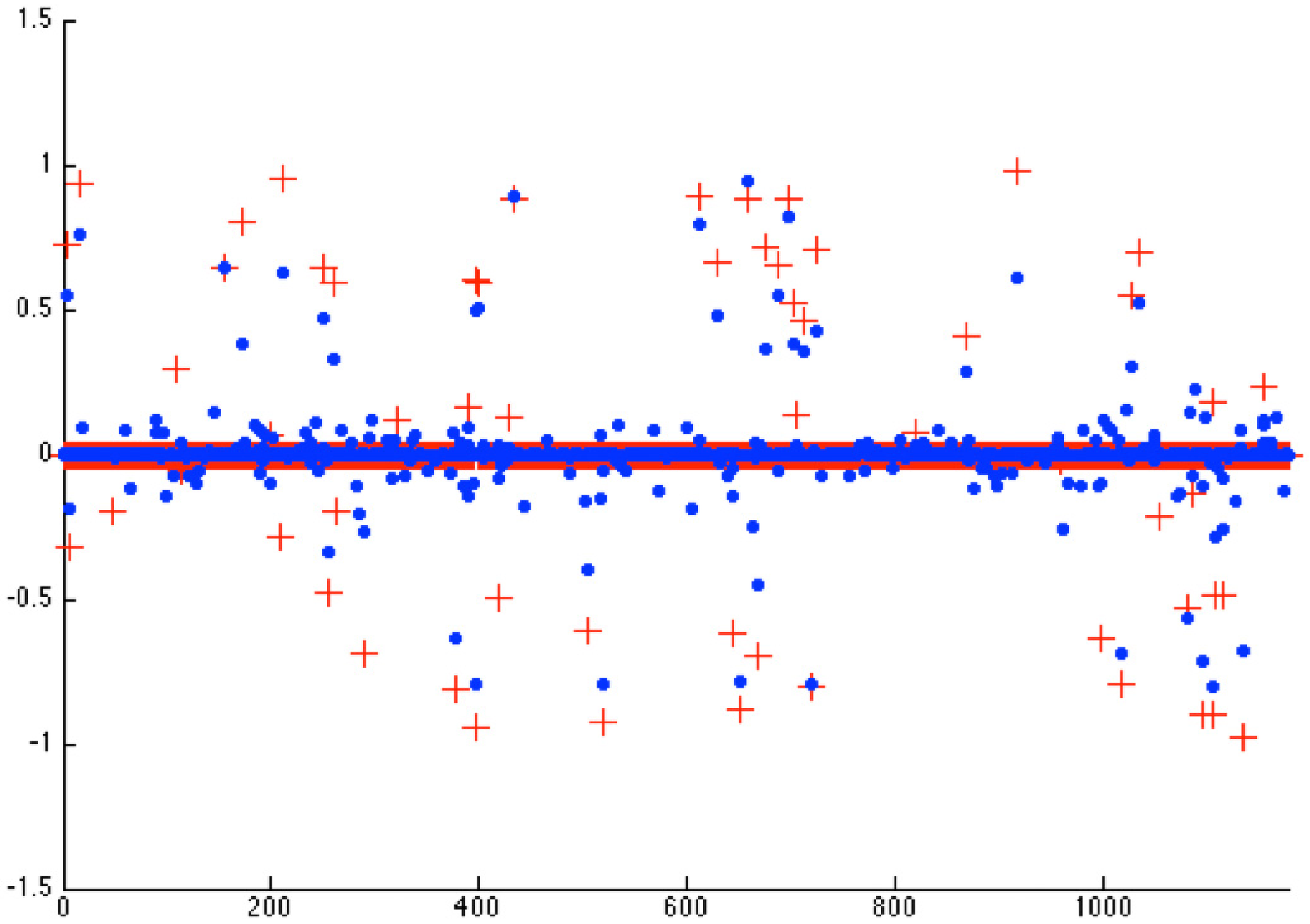}}
\hspace{0.2cm}
\subfigure[20NEWS, $\ell_1$]{\includegraphics[width=4.5cm]{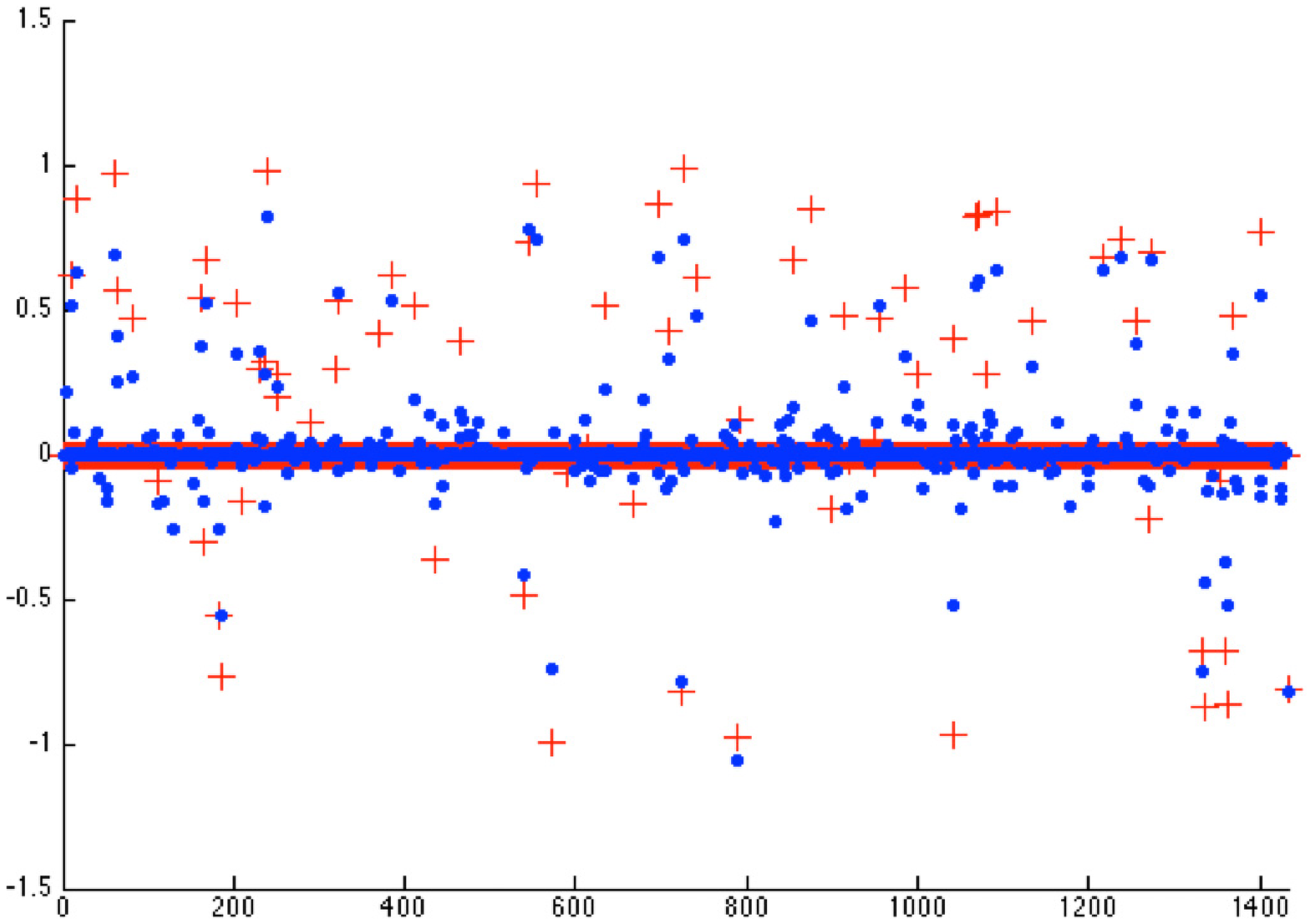}}\\
\subfigure[LFR, $\ell_1/ \ell_2$]{\includegraphics[width=4.5cm]{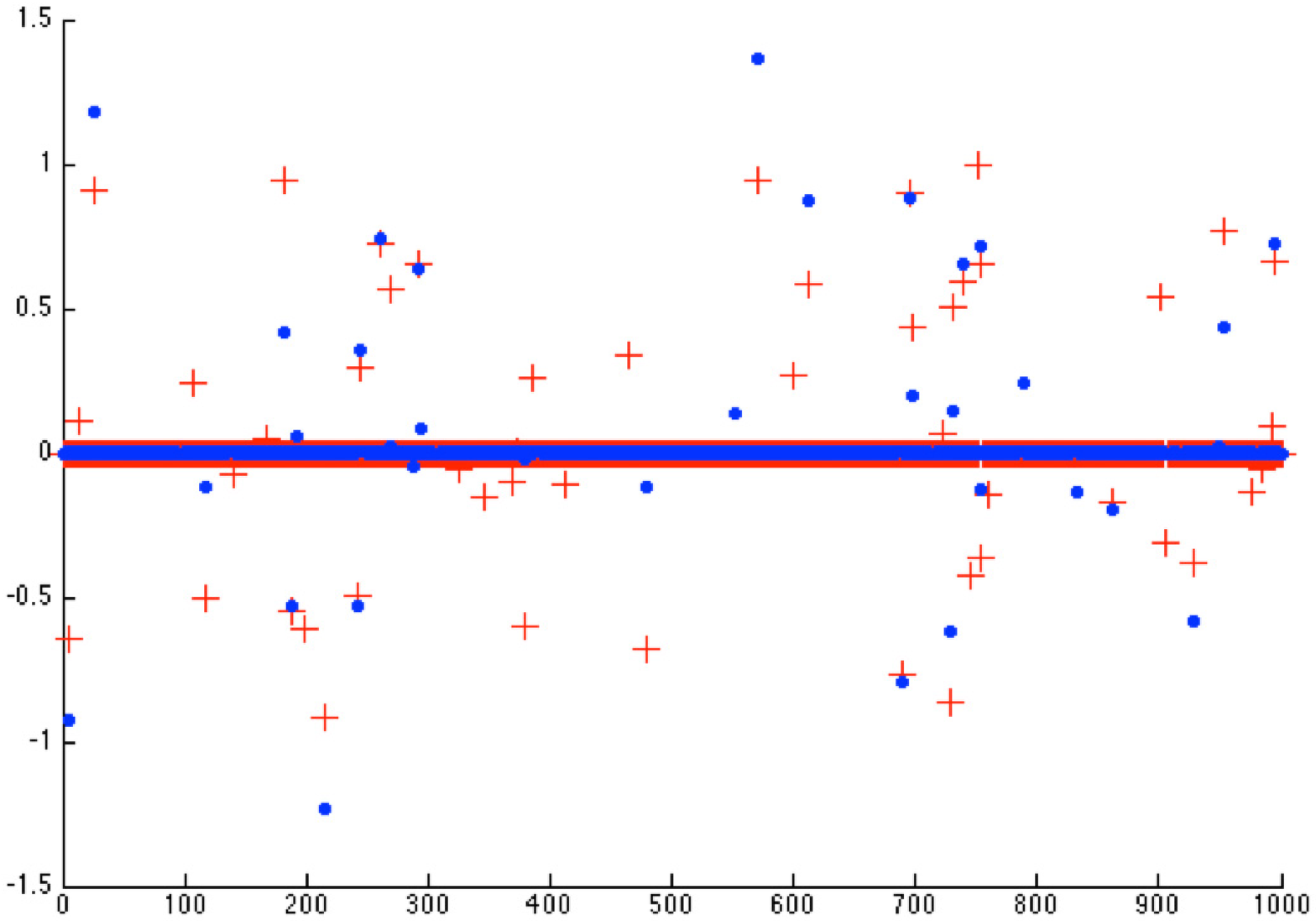}}
\hspace{0.2cm}
\subfigure[MNIST, $\ell_1/ \ell_2$]{\includegraphics[width=4.5cm]{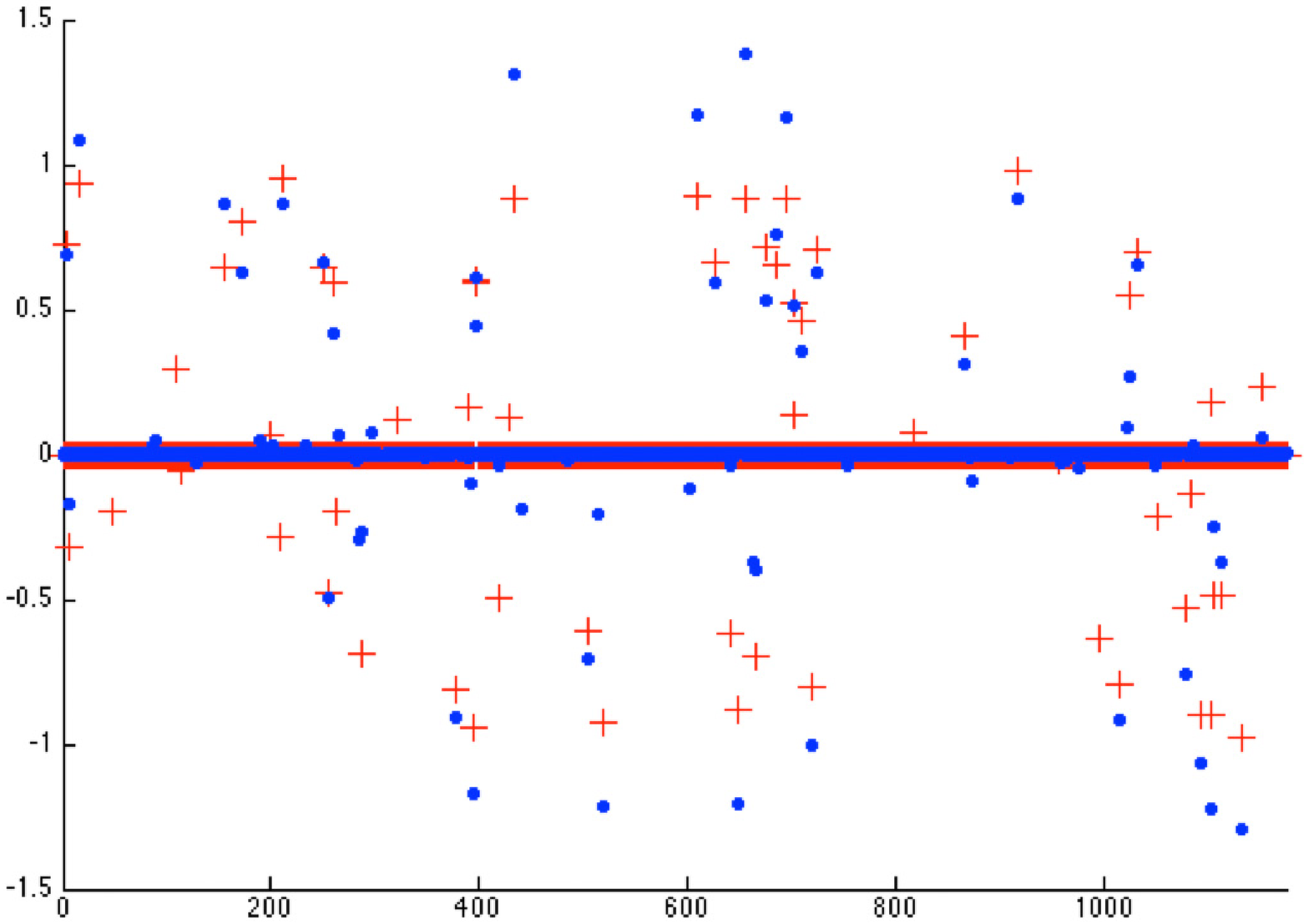}}
\hspace{0.2cm}
\subfigure[20NEWS, $\ell_1/ \ell_2$]{\includegraphics[width=4.5cm]{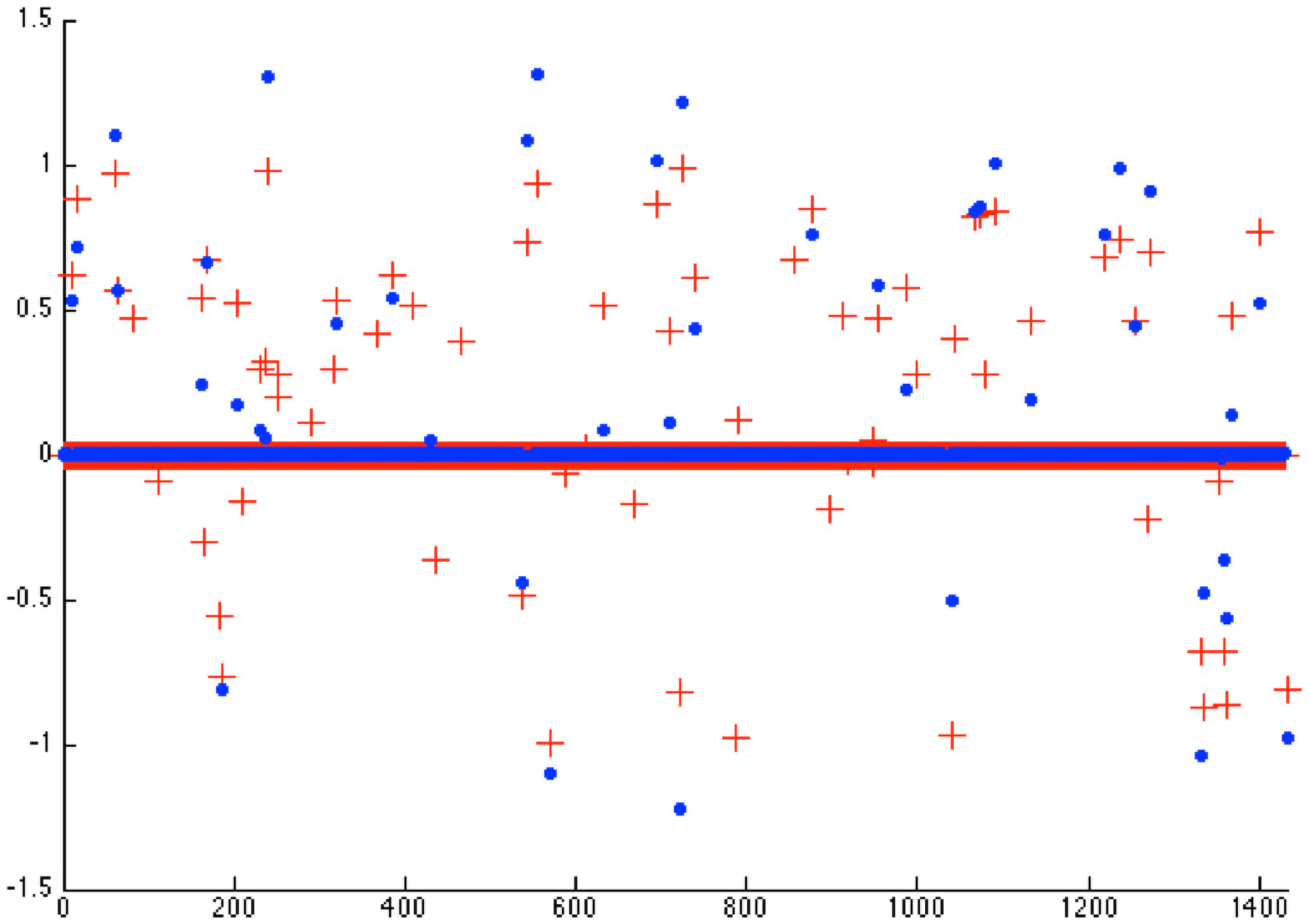}}
\caption{Standard Lasso-Inpainting vs Proposed Lasso-Inpainting on three graphs.}
\label{fig4}
\end{figure}

\noindent
{\bf Numerical Experiments.} We compare standard Lasso-Inpainting and enhanced Lasso-Inpainting on graphs. We test on the LFR, MNIST and 20NEWS graphs. We remove 40\% of measurements of $f_0$ with $R$. The value of the parameter $\lambda$ is again chosen to minimize the recovery error defined as $ \| x-x_0\|_2 / \| x\|_2$ for all models and all graphs. The results are reported on Table \ref{tab2} and Figure \ref{fig4}. Overall, the proposed enhanced Lasso-Inpainting model also performs better than the standard one, but it is 2-3 times slower.

\begin{table}[h!]
\centering
\begin{tabular}{|c|c|c|}
\hline
  & Standard Lasso-Inp  & Proposed Lasso-Inp \\
  \hline
 LFR & 0.667  & {\bf 0.540} \\
 MNIST & 0.509  & {\bf 0.362} \\
 20NEWS & 0.516  & {\bf 0.468} \\
 \hline	
\end{tabular}
\caption{Accuracy for standard Lasso-Inpainting vs proposed Lasso-Inpainting on three graphs.}
\label{tab2}
\end{table}

\section{Conclusion}
\label{sec5}

A new sparse recovery algorithm for Lasso-type problems on graph has been introduced. Numerical experiments have shown improvements over the standard $\ell_1$ algorithms. This result leverages the recent idea to go beyond $\ell_1$ convexity and explore non-convex, non-smooth techniques to find better sparse solutions. In this context, the closest works to ours are (i) the difference of convex (DC) functions \cite{art:LouOsherXin15DC} and (ii) the smoothed $\ell_1 / \ell_2$ technique \cite{art:RepettiQuyenDuvalChouzenouxPesquet15L1L2}. We would like to explore in a future work the relationship between our model and these models. Particularly, a direct application of Dinkelbach technique \cite{art:Dinkelbach67RatioOpt} reveals that minimizing the ratio is equivalent to minimize the DC model $\ell_1 - \alpha \ell_2$ with $\alpha$ being the minimum value of the ratio $\ell_1 / \ell_2$. As a result, an interesting question is whether this $\alpha$ value, which is automatically learned with the proposed algorithm, can provide satisfying solutions for a range of sparse problems. Eventually, we would like to compare our exact $\ell_1 / \ell_2$ ratio technique, which has a weak monotonicity property, with the smoothed ratio technique of \cite{art:RepettiQuyenDuvalChouzenouxPesquet15L1L2}, which has a strong monotonicity feature.\\

\noindent
{\bf Acknowledgements.} Thomas Laurent is supported by NSF grant DMS-1414396.




\bibliographystyle{unsrt}
\bibliography{bib_lasso_ratio}

\end{document}